\documentclass[11pt,a4paper]{article}
\usepackage{graphicx}
\usepackage[latin1]{inputenc}
\usepackage{amssymb,amsmath,array}
\usepackage{hyperref}
\usepackage{multirow}
\usepackage{subcaption}
\usepackage{array}
\usepackage{geometry}
\usepackage{authblk}

\DeclareMathOperator*{\argmax}{\arg\!\max}

\begin{document}
\title{Learning Sparse Feature Representations using Probabilistic Quadtrees and Deep Belief Nets}


\author[1]{Saikat Basu}
\author[1]{Manohar Karki}
\author[2]{Sangram Ganguly}
\author[1]{Robert DiBiano}
\author[1]{Supratik Mukhopadhyay}
\author[3]{Ramakrishna Nemani}
\affil[1]{Department of Computer Science, Louisiana State University}
\affil[2]{Bay Area Environmental Research Institute/ NASA Ames Research Center}
\affil[3]{NASA Advanced Supercomputing Division/ NASA Ames Research Center}

\renewcommand\Authands{ and }


\maketitle

\begin{abstract}
Learning sparse feature representations is a useful instrument for solving an unsupervised learning problem. In this paper, we present three labeled handwritten digit datasets, collectively called n-MNIST. Then, we propose a novel framework for the classification of handwritten digits that learns sparse representations using probabilistic quadtrees and Deep Belief Nets. On the MNIST and n-MNIST datasets, our framework shows promising results and significantly outperforms traditional Deep Belief Networks. 
\end{abstract}

\section{Introduction}
\emph{Deep Learning} has gained popularity over the last decade due to its ability to learn data representations in an unsupervised manner and generalize to unseen data samples using hierarchical representations. The most recent and best-known Deep learning model is the \emph{Deep Belief Network}\cite{Hinton06afast}. In \cite{MohamedDH12}, Deep Belief Networks have been used for modeling acoustic signals and have been shown to outperform traditional approaches using Gaussian Mixture Models for Automatic Speech Recognition (ASR). 
Deep Belief Network is trained one layer at a time using Restricted Boltzmann Machines (RBM). A sparse feature learning algorithm for Deep Belief Networks was proposed in \cite{RAN08}. However, their work was focused on maximization of information content in the learned representations. Restricted Boltzmann Machines, on the other hand, are trained by minimizing a contrastive term in the loss function.  

The main contributions of our work are twofold -- (1) We first present three labeled handwritten digit datasets, collectively called n-MNIST, created by adding white gaussian noise, motion blur and reduced contrast to the original MNIST dataset\cite{mnist}. (2) Then, we present a framework for the classification of handwritten digits that a) learns probabilistic quadtrees from the dataset, b) performs a Depth First Search on the quadtrees to create sparse representations in the form of linear vectors, and c) feeds the linear vectors into a Deep Belief Network for classification. On the MNIST and n-MNIST datasets, our framework shows promising results and significantly outperforms traditional Deep Belief Networks. 

\section[Datasets]{Datasets\footnote{The datasets are available at the web link \cite{datasets} along with a detailed description of the methods and parameters used to create them}}
\begin{figure}[h!]
\centering
\begin{subfigure}{.32\textwidth}
  \centering
\includegraphics[width=1\textwidth]{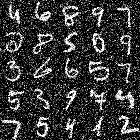}
\caption{MNIST with Additive White Gaussian Noise} 
\end{subfigure} 
\begin{subfigure}{.32\textwidth}
  \centering
\includegraphics[width=1\textwidth]{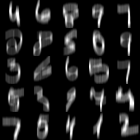}
\caption{MNIST with Motion Blur} 
\end{subfigure}  
\begin{subfigure}{.32\textwidth}
  \centering
\includegraphics[width=1\textwidth]{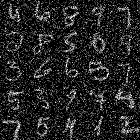}
\caption{MNIST with AWGN and reduced contrast} 
\end{subfigure} 
\caption{Example images from the n-MNIST dataset created as part of the experiments} \label{Fig:1}   
\end{figure}

We evaluate our framework on the MNIST dataset\cite{mnist} of handwritten digits as well as three artificial datasets collectively called n-MNIST (noisy MNIST) created by adding -- (1) additive white gaussian noise, (2) motion blur and (3) a combination of additive white gaussian noise and reduced contrast to the MNIST dataset. Some of the images from these datasets are shown in Figure \ref{Fig:1}. 

\section{Probabilistic Quadtrees for Learning Sparse Representations}
We propose a novel technique based on probabilistic quadtrees that can reduce the dimensionality of a dataset in a probabilistically sound way. We learn the structure of the quadtree from the samples of a dataset. A quadtree splits each image into four equi-sized windows, and then performs a test of homogeneity on each image window. If a block meets the homogeneity criterion, it is not divided any further into sub-windows. If otherwise, it does not meet the criterion, it is again subdivided into four sub-windows, and the test criterion is in turn applied to those smaller windows. This process is repeated on all the sub-windows until each meets the homogeneity criterion. The resulting data structure can have windows of several different sizes. The homogeneity criterion can be defined as follows - Split a block if the maximum value of the block elements minus the minimum value is greater than a threshold $\tau$. Threshold $\tau$ is specified as a value between $0$ and $1$ (chosen here as 0.27 by experiments). Denoting the homogeneity criterion for sample $d$ as $H_d$, this can be formally presented as follows:

\begin{equation}
H_d = 
\begin{cases}
true,
\hspace{5mm} \text{if} \max\limits_{i{\in}d} (i)-\min\limits_{i{\in}d}(i) \leq \tau \mid \tau\in[0,1] \\ 
false,
\hspace{3mm} \text{if} \max\limits_{i{\in}d} (i)-\min\limits_{i{\in}d}(i) > \tau \mid \tau\in[0,1] \\ 
\end{cases}
\end{equation}

Alternatively, the homogeneity criterion can be considered proportional to the standard deviation of the probability distribution of the dataset. So, higher the standard deviation, higher the average texture of a block and higher is the probability of the block being divided into sub-blocks.

In the learned quadtree structure for a given dataset, a node is divided into smaller windows if the homogeneity criterion is not met for any sample in the dataset.  The node is not divided into smaller windows only if the homogeneity criterion is met by all samples in the dataset.  

We can consider each node of the quadtree as a binary random variable $X$, which can take one of two values $1$ or $0$ based on whether it is divided into smaller windows or not. So, for a total of $N$ samples in dataset $D$, the random variable $X$ may take on one of $N + 1$ possible split state values: one value for each of the samples not meeting the homogeneity criterion, and one value indicating that all samples meet the homogeneity criterion. This can be formally presented as follows:

\begin{equation}
X = 
\begin{cases}
1,
\hspace{3mm} \text{if} ~\exists d \in D \mid D = \{d_0, d_1, d_2, ... , d_N\}  \cap  \{ H_d = false \} \\ 
0,
\hspace{3mm} \text{if} ~\forall d \in D \mid D = \{d_0, d_1, d_2, ... , d_N\}  \cap \{ H_d = true \} \\ 
\end{cases}
\end{equation}

Once learned, the probabilistic quadtree helps in reducing the dimensionality of the data, which captures the statistics of the training samples in the dataset. A depth first search on the learned tree yields a linear vector that is then fed into an unsupervised learning framework. 

\section{Deep Belief Network for Feature Learning} \label{sec:DBN}
\emph{Deep Belief Network} (DBN) consists of multiple layers of stochastic, latent variables trained using an unsupervised learning algorithm followed by a supervised learning phase using Feedforward Backpropagation Neural Networks.  In the unsupervised pre-training stage, each layer is trained using a \emph{Restricted Boltzmann Machine} (RBM). Once trained, the weights of the DBN are used to initialize the corresponding weights of a Neural Network \cite{Bengio2009}. A Neural Network initialized in this manner converges much faster than an otherwise uninitialized one. A DBN is a graphical model \cite{Koller2009} where neurons of the hidden layer are conditionally independent of each other given a particular configuration of the visible layer and vice versa. A DBN can be trained layer-wise by iteratively maximizing the conditional probability of the input vectors or visible vectors given the hidden vectors and a particular set of layer weights. As shown in \cite{Hinton06afast}, this layer-wise training can help in improving the variational lower bound on the probability of the input training data, which in turn leads to an improvement of the overall generative model. 
We first provide a formal introduction to the Restricted Boltzmann Machine.  The RBM can be denoted by the energy function:
\begin{equation}
E(v,h) = -\sum_{i} a_i v_i - \sum_{j} b_j h_j - \sum_{i} \sum_{j} h_j w_{i,j} v_i 
\end{equation}

where, the RBM consists of a matrix of layer weights $W=(w_{i,j})$ between the hidden units $h_j$ and the visible units $v_i$. The $a_i$ and $b_j$ are the bias weights for the visible units and the hidden units respectively.
The RBM takes the structure of a bipartite graph and hence it only has inter-layer connections between the hidden or visible layer neurons but no intra-layer connections within the hidden or visible layers. So, the visible unit activations are mutually independent given a particular set of hidden unit activations and vice versa \cite{carreiraperpinan2005contrastive}.  Hence, by setting either $h$ or $v$ constant, we can compute the conditional distribution of the other as follows:

\begin{equation}
P(h_j=1|v) = \sigma(b_j + \sum_{i=1}^{m} w_{i,j} v_{i})
\end{equation}

\begin{equation}
P(v_i=1|h) = \sigma(a_i + \sum_{j=1}^{n} w_{i,j} h_{j})
\end{equation}

where, $\sigma$ denotes the log sigmoid function:

\begin{equation}
f(x) = \frac{1}{1+e^{-x}}
\end{equation}

The training algorithm maximizes the expected log probability assigned to the training dataset $D$. So if the training dataset $D$ consists of the visible vectors $v$, then the objective function is as follows:

\begin{equation}
\argmax_{W} E\Big[\sum_{v \in V} \log{P(v)} \Big]
\end{equation}

A Restricted Boltzmann Machine is trained using a \emph{Contrastive Divergence} algorithm \cite{carreiraperpinan2005contrastive}. Once trained the DBN is used to initialize the weights of a feedforward backpropagation neural network that is then used for classification.
\begin{table}[h!]
\centering
\begin{tabular}{ | c | c | c | c | c |}
    \hline
    & \multicolumn{2}{|c|}{\textbf{MNIST}} & \multicolumn{2}{|c|}{\textbf{n-MNIST with AWGN}}\\ \hline
    \textbf{Architecture} & \textbf{Test Error} & \textbf{Test Error} & \textbf{Test Error} & \textbf{Test Error} \\ 
    \textbf{(Neurons)} & \textbf{DBN(\%)} & \textbf{Ours(\%)} & \textbf{DBN(\%)} & \textbf{Ours(\%)} \\ \hline
    50-50 & 4.64 & 2.93 & 89.95 & 13.41 \\ \hline
    100-100 & 3.01 & 2.45 & 91.43 & 12.01 \\ \hline
    150-150 & 2.34 & 2.21 & 89.95 & 13.49 \\ \hline
    200-200 & 2.08 & 1.96 & 88.49 & 10.56 \\ \hline
    250-250 & 1.93 & 1.83 & 88.49 & 13.00 \\ \hline
    300-300 & 2.02 & 1.80 & 68.18 & 11.24 \\ \hline
    350-350 & 1.96 & 1.74 & 90.31 & 13.15 \\ \hline
    400-400 & 1.95 & 1.67 & 49.27 & 10.96 \\ \hline
    450-450 & 1.93 & \textbf{1.38} & \textbf{32.26} & 12.62 \\ \hline
    500-500 & \textbf{1.86} & 1.43 & 69.68 & \textbf{9.93} \\ \hline
  \end{tabular}
  \caption{Test Error of a traditional DBN and our framework with various architectures on MNIST and n-MNIST with AWGN}
  \label{table:error_comparison_1}
\end{table}

\begin{table}[h!]
\centering
\begin{tabular}{ | c | c | c | c | c |}
    \hline
    & \multicolumn{2}{|c|}{\textbf{n-MNIST with}} & \multicolumn{2}{|c|}{\textbf{n-MNIST with AWGN}}\\ 
    & \multicolumn{2}{|c|}{\textbf{Motion Blur}} & \multicolumn{2}{|c|}{\textbf{and Reduced Contrast}}\\ \hline
    \textbf{Architecture} & \textbf{Test Error} & \textbf{Test Error} & \textbf{Test Error} & \textbf{Test Error} \\ 
    \textbf{(Neurons) } & \textbf{DBN(\%)} & \textbf{Ours(\%)} & \textbf{DBN(\%)} & \textbf{Ours(\%)} \\ \hline
    50-50 & 5.64 & 4.17 & 10.21 & 9.29 \\ \hline
    100-100 & 4.68 & 3.31 & \textbf{9.43} & 9.21 \\ \hline
    150-150 & 3.99 & 3.29 & 16.40 & 9.00 \\ \hline
    200-200 & 3.74 & 3.03 & 15.57 & 8.79 \\ \hline
    250-250 & 3.74 & \textbf{2.60} & 52.31 & 8.94 \\ \hline
    300-300 & \textbf{3.50} & 3.04 & 32.29 & 8.28 \\ \hline
    350-350 & 3.82 & 2.91 & 86.31 & 8.90 \\ \hline
    400-400 & 3.74 & 3.01 & 68.78 & 8.31 \\ \hline
    450-450 & 3.91 & 2.75 & 51.32 & 8.36 \\ \hline
    500-500 & 3.66 & 2.83 & 68.19 & \textbf{7.84} \\ \hline
  \end{tabular}
  \caption{Test Error of a traditional DBN and our framework with various architectures on n-MNIST with Motion Blur; and with AWGN and Reduced Contrast}
  \label{table:error_comparison_2}
\end{table}

\section{Results and Comparative Studies}
Various network architectures along with the test set error for the traditional DBN framework and the probabilistic quadtree based framework on the MNIST and the three n-MNIST datasets are listed in Tables \ref{table:error_comparison_1} and \ref{table:error_comparison_2}. From the Tables, it is evident that our best performing network outperforms the best traditional Deep Belief Network on both the MNIST and n-MNIST datasets. On the MNIST dataset, our best network exhibits a relative improvement of ${\sim}$25\% over the traditional DBN. For the n-MNIST dataset, it provides a relative improvement of ${\sim}$36\% for Additive White Gaussian Noise (AWGN), ${\sim}$26\% for Motion Blur and ${\sim}$12\% for AWGN and Reduced contrast.

\section{Discussion and Future Directions}
Our learning framework based on probabilistic quadtrees significantly outperforms traditional Deep Belief Networks on both the MNIST and n-MNIST datasets. Probabilistic quadtrees help in generating sparse representations for the dataset and significantly improve the discriminative power of the framework. 

We plan to investigate the use of various pooling techniques like SPM \cite{Lazebnik:2006} as well as certain sparse representations like sparse coding \cite{Lee07efficientsparse} to handle n-MNIST. Hierarchical representations like Convolutional DBN \cite{Lee:2009:CDBN} are other useful candidates for investigation. We believe that n-MNIST will help researchers better apply and extend the research on understanding representations for noisy object recognition datasets.

\section*{Acknowledgment}

The project is supported by Army Research Office (ARO) under Grant \#W911-NF1010495 and NASA Carbon Monitoring System through Grant \#NNH14ZD-A001NCMS. Any opinions, findings, and conclusions or recommendations expressed in this material are those of the authors and do not necessarily reflect the views of the ARO or the United States Government.


\bibliographystyle{unsrt}
\bibliography{References}


\end{document}